\documentclass{article}
\usepackage{spconf,amsmath,epsfig}
\usepackage{multirow}

\title{
Graph Neural Network as Computationally Efficient Emulator of Ice-sheet and Sea-level System Model (ISSM)}
\name{Younghyun Koo, Maryam Rahnemoonfar}
\address{Department of Computer Science and Engineering\\
Department of Civil and Environmental  Engineering\\
Lehigh University\\
Bethlehem, PA 18015, USA}
\begin{document}
%
\maketitle

\begin{abstract}
The Ice-sheet and Sea-level System Model (ISSM) provides solutions for Stokes equations relevant to ice sheet dynamics by employing finite element and fine mesh adaption. However, since its finite element method is compatible only with Central Processing Units (CPU), the ISSM has limits on further economizing computational time. Thus, by taking advantage of Graphics Processing Units (GPUs), we design a graph convolutional network (GCN) as a fast emulator for ISSM. The GCN is trained and tested using the 20-year transient ISSM simulations in the Pine Island Glacier (PIG). The GCN reproduces ice thickness and velocity with a correlation coefficient greater than 0.998, outperforming the traditional convolutional neural network (CNN). Additionally, GCN shows 34 times faster computational speed than the CPU-based ISSM modeling. The GPU-based GCN emulator allows us to predict how the PIG will change in the future under different melting rate scenarios with high fidelity and much faster computational time.
\end{abstract}

\section{Introduction}\label{sec:intro}
Recently, as the global climate has been warming, the Antarctic ice sheets have lost around 2,720 billion tonnes of ice mass since 1992, which corresponds to an increase in mean sea level
of 7.6 mm \cite{IMBIE}. In particular, the Pine Island Glacier (PIG) has shown the most rapid mass loss and speedups, contributing to more than 20 \% of the total sea level rise in Antarctica \cite{Rignot2019}. Among various environmental factors, melt-driven thinning near the grounding line and calving events have been blamed for driving such changes in the PIG \cite{Joughin2021, Joughin2021_2, jacobs2011}. Therefore, in order to predict the future mass changes of the PIG and its impact on the rise of global sea level, it is necessary to model the thermodynamic and dynamic reaction of the PIG to external heat sources (e.g., ocean warming).

To explain the dynamics of ice sheets, scientists have assumed ice is viscous and non-Newton fluid governed by the Stokes equation \cite{Glen1955}. Based on the Stokes equation, the flow of large ice sheets can be described by several simplified approximations: e.g., Shallow Ice Approximation (SIA) \cite{Hutter1983}, Shallow Shelf Approximation (SSA) \cite{MacAyeal1989}, Blatter-Pattyn approximation (BP) \cite{Blatter1995, Pattyn1996}, and a full-Stokes model (FS) \cite{Morlighem2010}. The Ice-sheet and Sea-level System Model (ISSM) provides numerical solutions for these four different ice flow models using a finite element approach \cite{Larour2012}. The ISSM has significant advantages in fast and accurate ice modeling for the following reasons: (1) finite element methods, (2) fine mesh adaptation, and (3) parallel technologies. The unstructured meshes and anisotropic mesh refinement allow ISSM to allocate its computational resources to the fine-resolution area of fast ice and coarse-resolution areas of stagnant ice. Additionally, state-of-the-art parallel technologies reduce the running time of the ISSM model dramatically when implemented in massive computer clusters \cite{Larour2012}.

Although the parallel technologies of the ISSM allow for saving computational time, its computational performance is still limited because it is only available through multi-core central processing units (CPUs). Recently, graphic processing units (GPUs) have emerged as an attractive alternative processor to CPUs due to their specialty in parallel processing. GPUs divide a task into several small tasks and process them parallelly, which allows a magnificent speed-up compared to serial processing by CPU. Nevertheless, considering that every finite element has to be solved one at a time in ice flow modeling, the parallel computation of GPUs cannot be directly used to solve the governing equations of ice flow. Hence, instead of using GPU to solve the ice flow equations directly, a GPU-based machine learning model can be used as a statistical emulator (a.k.a. surrogate model) to accelerate the computational time on ice sheet modeling \cite{Jouvet2022, jouvet2023, jouvet2021_inversion, He2023}.

In this study, we develop a new deep-learning emulator of the ISSM by selecting the PIG as the test site. To take full advantage of the finite element and mesh adaption of the ISSM, we employ a graph neural network (GNN) instead of the traditional convolutional neural network (CNN) that has been used in most previous studies \cite{Jouvet2022, jouvet2023, jouvet2021_inversion}. Although traditional CNNs succeed in recognizing spatial patterns of Euclidean or grid-like structures (e.g., images) by using fixed-size trainable localized filters, CNNs cannot be used for non-Euclidean or irregular structures where the connections to neighbors are not fixed \cite{zhang2019}. Hence, for irregular structures, including point clouds, natural languages, molecular structures, and social networks, GNNs have been more broadly used \cite{zhang2019}. Since the meshes of the ISSM model resemble irregular graph structures, GNN is more appropriate than other deep-learning architectures in reproducing the ISSM model. In this study, by substituting the mesh defined in ISSM with a graph structure, we focus on developing and testing a graph convolutional network (GCN) that imitates the ability of the ISSM to model ice flow and thickness. Furthermore, by using this fast GCN emulator, we examine how the ice flow and thickness of the PIG will change by ocean warming and ice melting.

\section{Methods}

\subsection{ISSM simulation for the Pine Island Glacier}

\begin{figure}
    \centering
    \includegraphics[width=1.0\linewidth]{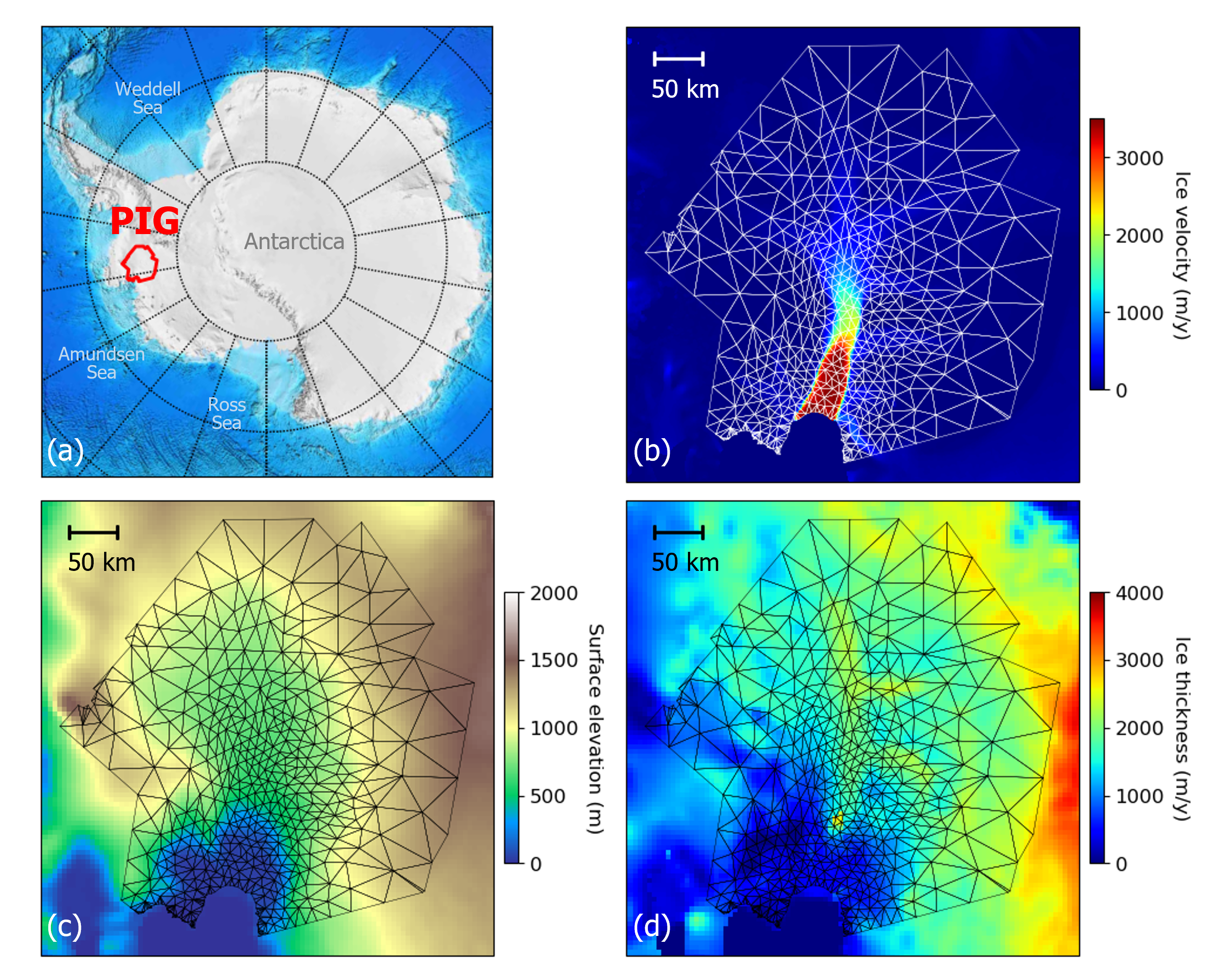}
    \caption{(a) Location of the Pine Island Glacier (PIG) in Antarctica; (b) Ice velocity, (c) surface elevation, and (d) ice thickness of the PIG acquired from NASA MEaSUREs data.}
    \label{PIG}
\end{figure}

We simulate the 20-year evolution of ice thickness and ice velocity in the PIG by following the previous studies of ISSM-based sensitivity experiments \cite{Seroussi2014_PIG, Laour2012_PIG}. For ice velocity, thickness, surface elevation, and bed topography data, we use the NASA Making Earth System Data Records for Use in Research Environments (MEaSUREs) data \cite{MEASURE_thickness, MEASURE_velocity} (Fig.\ref{PIG}). SSA \cite{MacAyeal1989} is used as an ice flow model with 10 km of initial mesh sizes ($M_0$). The final mesh consists of 2085 elements and 1112 nodes. To check the impact of basal melting rate on ice thickness and ice velocity, we change annual basal melting rates from 0 to 70 m/year for every 2 m/year. This basal melting rate is only applied for the floating ice; ground ice is assumed to have zero melting rate. Transient simulations are run forward for 20 years with time steps of one month, and all results for 240 months and 36 melting rates are collected.

\subsection{Graph Convolutional Network (GCN)}

\begin{figure}
    \centering
    \includegraphics[width=1.0\linewidth]{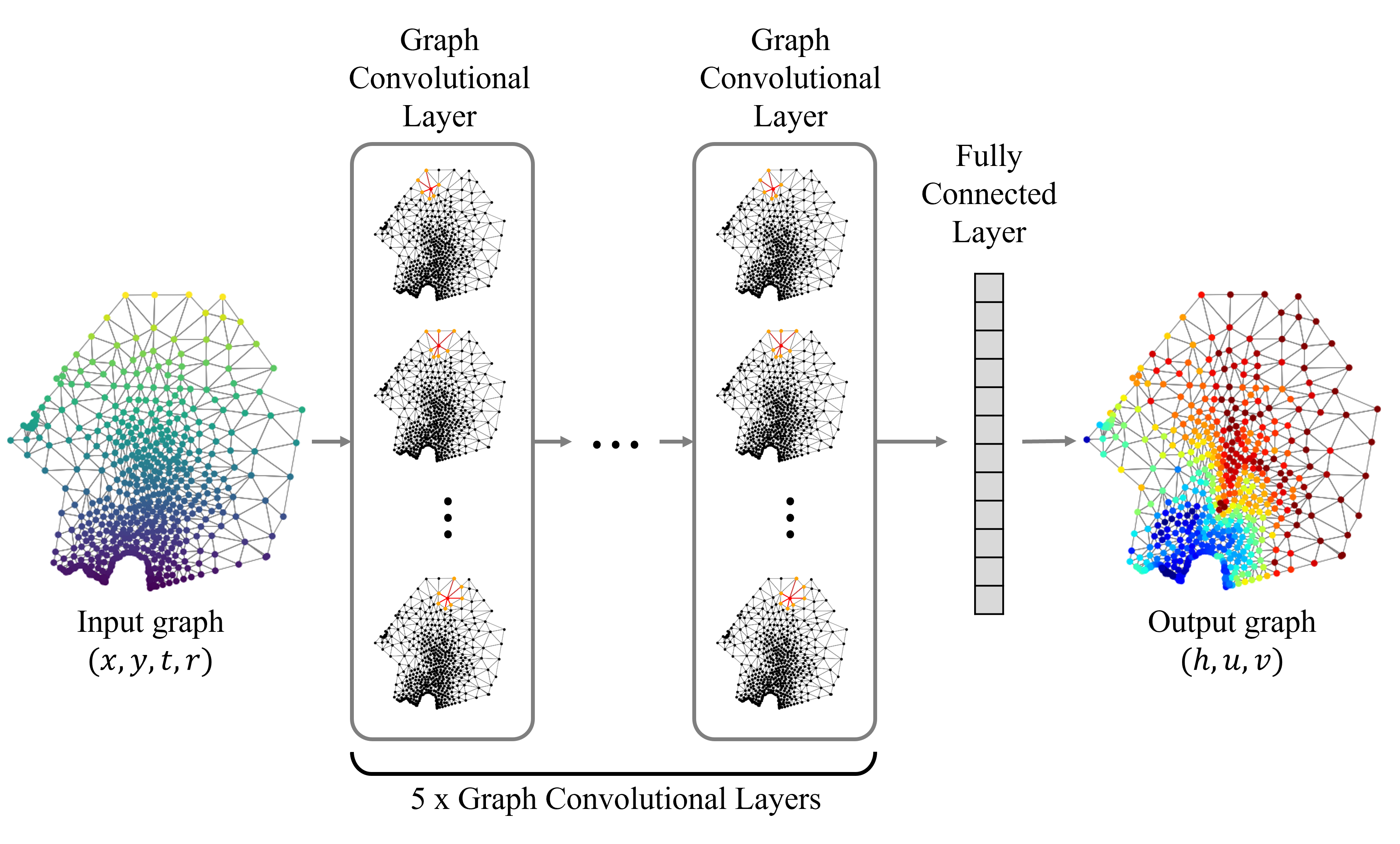}
    \caption{Schematic illustration of the graph convolutional network (GCN) emulator.}
    \label{gcn}
\end{figure}

The GCN of this study is based on the graph convolutional layer, which is inspired by a localized first-order approximation of spectral graph convolutions on graph-structured data \cite{Kipf2016_GCN}. 
Let the $l$th graph convolutional layer receives a set of node features $\textbf{h}^{(l)} = \{h_1^{(l)}, h_2^{(l)}, ..., h_N^{(l)}\}$, $h_i^{(l)} \in {R}^{F_{l}}$ as the input and produces a new set of node features,  $\textbf{h}^{(l+1)} = \{h_1^{(l+1)}, h_2^{(l+1)}, ..., h_N^{(l+1)}\}$, $h_i^{(l+1)} \in {R}^{F_{l+1}}$, for the next $l+1$th layer. $N$ is the number of nodes; $F_{l}$ and $F_{l+1}$ is the number of features in each node at $l$th layer and $l+1$ layer, respectively. Then, the layer-wise propagation rule of the graph convolutional layer is defined by the following equation:

\begin{equation}
\label{eq_gcn}
\begin{split}
    & h_i^{(l+1)}=\sigma(\sum_{j\in\mathcal{N}(i)}\frac{e_{ij}}{c_{ij}}\textbf{W}^{(l)}h_j^{(l)}) \\
    & e_{ij}=\text{exp}(-\frac{1}{\sqrt{(x_i-x_j)^2+(y_i-y_j)^2}})
\end{split}
\end{equation}
where $\mathcal{N}(i)$ is the set of neighbors of node $i$, $c_{ij}$ is the product of the square root of node degrees (i.e., $c_{ij}=\sqrt{|\mathcal{N}(j|)}\sqrt{|\mathcal{N}(i)|}$), $e_{ij}$ is the scalar weight determined by the spatial distance between node $i$ and node $j$. The x, y coordinates and node $i$ and node $j$ are denoted by $(x_i, y_i)$ and $(x_j, y_j)$, respectively. $\textbf{W}^{(l)}$ is a layer-specific trainable weight matrix ($\textbf{W}^{(l)} \in {R}^{F_{l+1} \times F_{l}}$), and $\sigma(\cdot)$ is an activation function; we use the Leaky ReLU activation function of 0.01 negative slope in this study. The GCN takes four input features (i.e., x and y coordinates on the Antarctic Polar Stereographic projection, time, and basal melting rate) and produces three output features (i.e., x-component and y-component ice velocity and ice thickness). We use five graph convolutional layers with 128 feature sizes and add a fully connected layer after the final graph convolutional layer to produce the outputs (Fig. \ref{gcn}).

\subsection{Fully Convolutional Network (FCN)}

We also train and test a fully convolutional network (FCN) as a baseline machine learning model to compare with GCN. FCN, a type of CNN, consists of only convolutional layers. In this study, we use an FCN with six convolutional layers; all the graph convolutional layers and fully connected layers in Fig. \ref{gcn} are replaced with 2-D convolutional layers with a kernel size of 3 and filter size of 128. The leaky ReLu activation function of 0.01 negative slope is applied after each convolutional layer. Since the FCN should take regular grids as the input and output, we interpolate all the irregular mesh construction of the ISSM simulation into a 1 km grid, and these 1 km grid datasets are used as the input and output of the FCN. The output of the FCN is sampled again to the ISSM mesh to be compared with the GCN output.

\section{Results}

\subsection{Experiment set-up}

We collect a total of 8,640 graph structures ($240 \text{ months} \times 36 \text{ basal melting rates}$) from the ISSM transient simulations. All 8,640 graph structures are divided into train, validation, and test datasets based on the melting rate values: melting rates of 10, 30, 50, and 70 m/year are used for validation, melting rates of 0, 20, 40, and 60 m/year are used for testing, and the remainders are for training. Consequently, the number of train, validation, and test datasets is 6,720, 960, and 960, respectively. We use the mean square error (MSE) as the loss function, and the model is optimized by Adam stochastic gradient descent algorithm with 200 epochs and 0.01 learning rate. All deep learning models are trained on the Python environment using the Deep Graph Library (DGL) and PyTorch modules. In measuring the computational time, we record the time to generate the final results of 20-year ice thickness and velocity for all 36 melting rates after training. Two computational resources of the same desktop (Lenovo Legion T5 26IOB6) are compared: a CPU (Intel(R) Core(TM) i7-11700F) and a GPU (NVIDIA RTX A5000).

\subsection{Fidelity of the graph neural network emulators}

\begin{figure}
    \centering
    \includegraphics[width=1.0\linewidth]{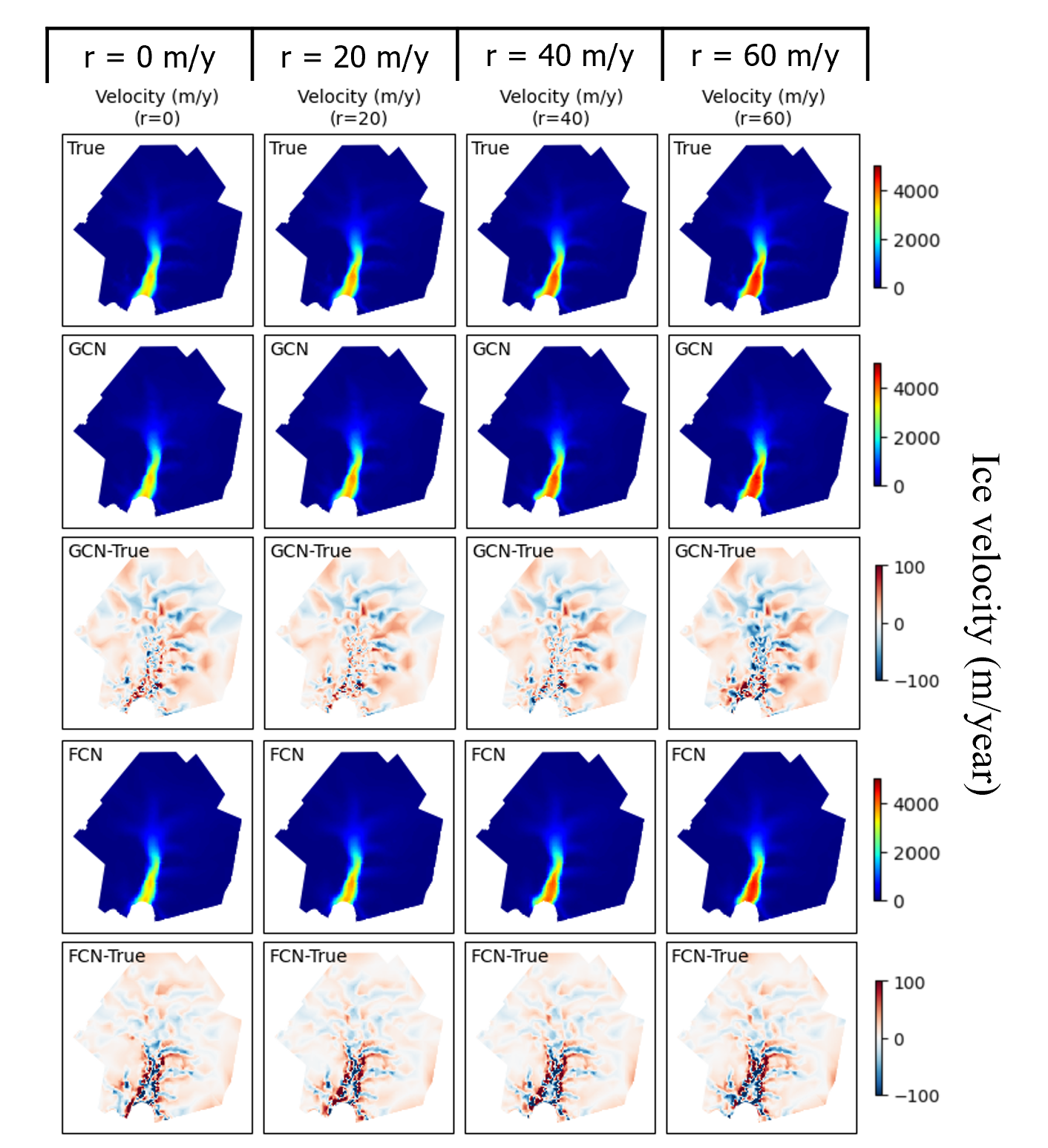}
    \caption{Maps of the 20-year-averaged ice velocity for different basal melting rates}
    \label{map_velocity}
\end{figure}

\begin{figure}
    \centering
    \includegraphics[width=1.0\linewidth]{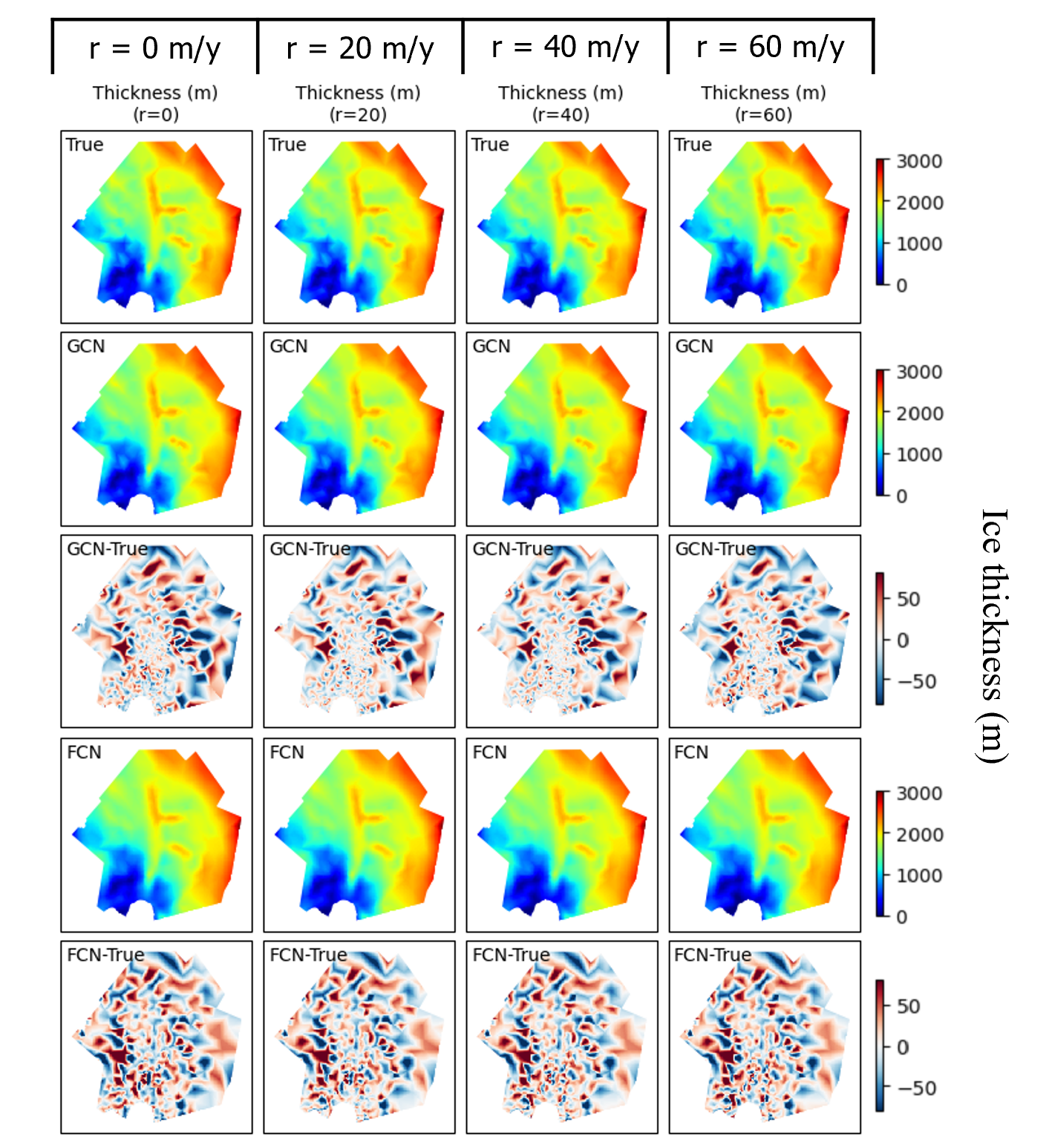}
    \caption{Maps of the 20-year-averaged ice thickness for different basal melting rates}
    \label{map_thickness}
\end{figure}

The accuracy of GCN and FCN machine learning surrogate models are described in Table \ref{table_fidelity}. Both GCN and FCN show high correlation coefficients greater than 0.996 with the ISSM results for ice velocity and thickness modeling. However, GCN generally shows better agreement with the ISSM results than the FCN. The RMSE of ice velocity from GCN is less than that from FCN by 42.55 m/year, and the RMSE of ice thickness from GCN is less than FCN by 9.64 m. This result shows that the graph construction and GCN architecture can represent the mesh structure of the ISSM better than the FCN (CNN). In other words, the fixed kernel (window) size of FCN cannot efficiently reproduce different neighboring impacts by different ice velocity fields. Indeed, as shown in Fig. \ref{map_velocity}, the error of FCN concentrates on the fast-ice area in the southern center of the PIG. Additionally, the ice thickness error of GCN is lower than FCN in this fast-ice area (Fig. \ref{map_thickness}).

\begin{table}
\caption{Accuracy of ice velocity and thickness for machine learning emulators.}
\begin{tabular}[t]{c|c|c|c|c}
\hline
\multirow{2}{*}{Model} &
\multicolumn{2}{c|}{Ice velocity} &
\multicolumn{2}{c}{Ice thickness}\\
\cline{2-5}
&RMSE (m/year) &R &RMSE (m) &R \\
\hline
GCN &\textbf{55.31} &\textbf{0.999} &\textbf{42.95} &\textbf{0.998}\\
FCN &97.86 &0.996 &52.59 &0.997\\
\hline
\end{tabular}
\label{table_fidelity}
\end{table}

\subsection{Computational performance of the graph neural network emulators}

GCN also shows a better computational performance than FCN (Table \ref{table_time}). When GPUs are used for machine learning models, GCN shows approximately 33.8 times faster computation time than ISSM computational time, which is only compatible with CPUs. On the other hand, FCN shows approximately 12.0 times faster computation time than ISSM. This result shows that both GCN and FCN based on GPUs can save significant time in obtaining transient simulation results, but GCN is more efficient.

\begin{table}
\caption{Computational time (in seconds) for producing final ice sheet modeling results. The percentage in the parentheses represents the percentage ratio to the ISSM simulation time.}
\begin{tabular}[t]{c|c|c|c}
\hline
\multirow{2}{*}{} &
\multirow{2}{*}{Model} &
\multicolumn{2}{c}{Computational time (seconds)}
\\\cline{3-4}
& &CPU &GPU \\
\hline
Simulation & ISSM & 399.78 & - \\
\hline
\multirow{2}{*}{Emulator} & GCN & 60.37 & 11.83  \\\cline{2-4}
 & FCN &1859.65 &33.27 \\\cline{2-4}
\hline
\end{tabular}
\label{table_time}
\end{table}

\subsection{Sensitivity to basal melting rate}

\begin{figure}
    \centering
    \includegraphics[width=1.0\linewidth]{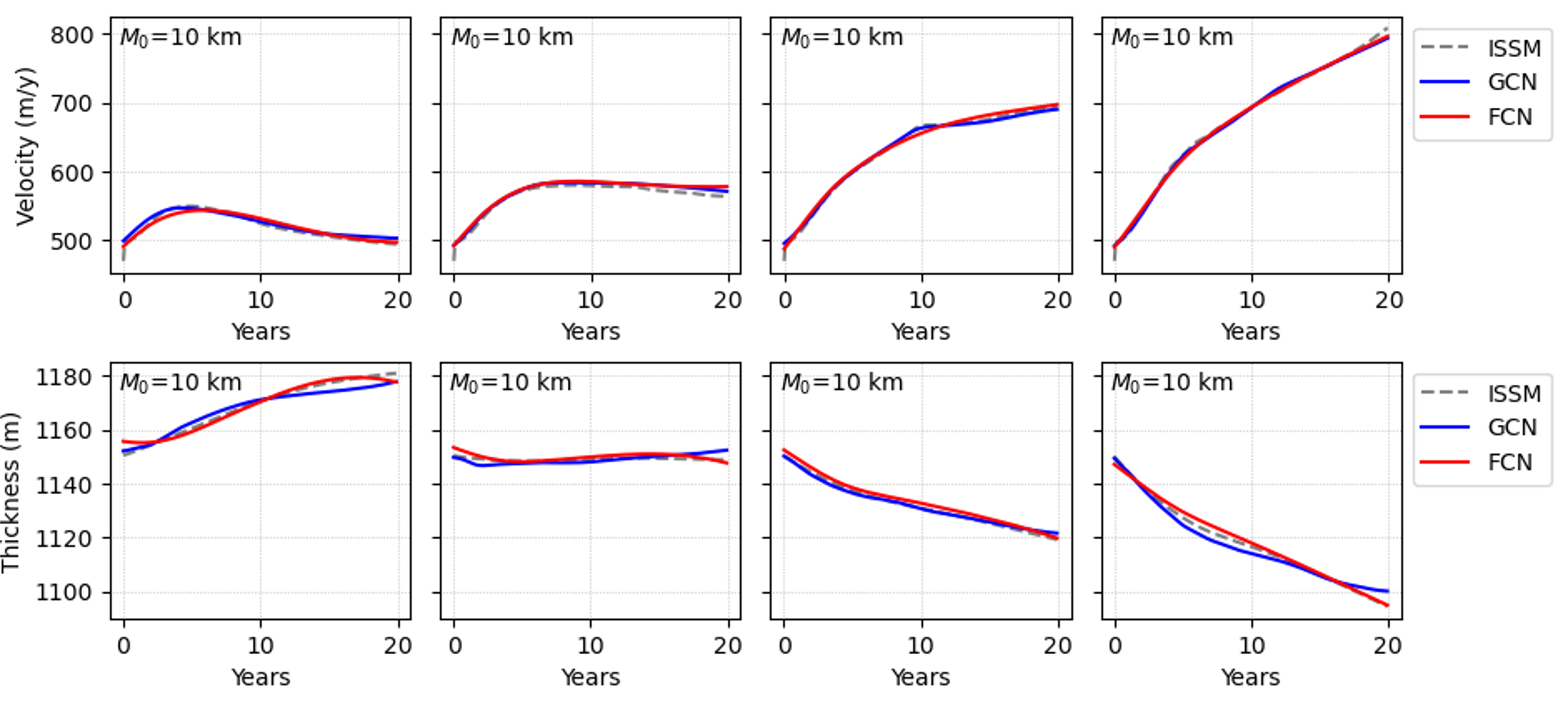}
    \caption{Changes in mean ice velocity and ice thickness of the PIG under different melting rates (0, 20, 40, and 60 m/year).}
    \label{Sensitivity}
\end{figure}

Finally, by using the GCN and FCN emulators, we examine how ice velocity and thickness of the PIG are sensitive to changes in basal melting rates. Fig. \ref{Sensitivity} shows the 20-year variations in the mean ice velocity and thickness modeled by ISSM, GCN, and FCN under four different melting rates (0, 20, 40, and 60 m/year). In all modeling results, ice velocity and thickness vary dramatically by different basal melting rates. Interestingly, no significant difference between GCN and FCN is observed, even though GCN shows better accuracy in ice velocity and thickness predictions. Under zero melting rate (i.e., no melting on floating ice), ice velocity remains relatively consistent over 20 years within 500-550 m/year of speed. However, under positive melting rates, ice velocity increases with time; the slope becomes steeper with a higher melting rate. If the melting rate is 60 m/year, the terminal ice velocity in year 20 reaches around 800 m/year, corresponding to almost 200 m/year of speed-up over 20 years. Such a dramatic increase in ice velocity can result in extreme calving at the ice front and large mass loss. When it comes to ice thickness, the mean ice thickness increases by 20-30 m over 20 years under zero melting rate. If the melting rate is 20 m/year, the mean ice thickness remains static over 20 years. However, if the melting rate exceeds 20 m/year, the mean ice thickness decreases continuously over 20 years, and this decreasing trend is accelerated at a higher melting rate. The decrease in mean ice thickness is expected to reach 50 m under an extremely high melting rate of 60 m/year. Under this 60 m/year ocean melting scenario, the PIG will lose approximately 1,800 Gt of ice for 20 years, equivalent to 2.57 mm of global sea level rise.

\section{Conclusion}

In this study, we assess the potential of graph convolutional network (GCN) as a fast and efficient emulator (surrogate model) for the ISSM (Ice-sheet and Sea-level System Model) numerical model. Compared to the traditional convolutional neural network (CNN), GCN shows better accuracy and computational efficiency in reproducing the ISSM modeling based on refined finite element construction. This GCN emulator is used for detailed sensitivity analysis about how the ice velocity and thickness in the PIG will change under different basal melting rate scenarios. The sensitivity analysis results exhibit that the mass loss and speed-up in the PIG will be dramatically accelerated under $>$ 20 m/year of melting rates. As a fast and high-fidelity emulator of the ISSM, GCN can be further used to investigate the impact of climate change and ocean warming on Greenland and Antarctic ice sheets.


\bibliographystyle{IEEEbib}
\bibliography{refs}

\end{document}